\documentclass{article}

\usepackage{arxiv}

\usepackage[utf8]{inputenc} % allow utf-8 input
\usepackage[T1]{fontenc}    % use 8-bit T1 fonts
\usepackage{hyperref}       % hyperlinks
\usepackage{url}            % simple URL typesetting
\usepackage{booktabs}       % professional-quality tables
\usepackage{amsfonts}       % blackboard math symbols
\usepackage{nicefrac}       % compact symbols for 1/2, etc.
\usepackage{microtype}      % microtypography
\usepackage{lipsum}		% Can be removed after putting your text content
\usepackage{graphicx}
\usepackage[numbers]{natbib}
\usepackage{doi}
\usepackage{amsmath}
\usepackage{booktabs} % For professional table formatting
\usepackage{tabularx} % For dynamic column width
\title{An efficient approach to represent enterprise web application structure using Large Language Model in the service of intelligent quality engineering}

\author{ {Zaber Al Hassan Ayon}
\thanks{https://www.linkedin.com/in/zaber-al-hassan-ayon-407290106} 
\\
	Atalgo Engineering\\
	Atalgo Computing Pty Ltd\\
	Perth, WA, 6000 \\
	\texttt{ayon@atalgo.com} \\
	\And
    {Gulam Husain}\thanks{https://www.linkedin.com/in/gulam-husain/} \\
	Atalgo Engineering\\
	Atalgo Computing Pty Ltd\\
	Perth, WA, 6000 \\
	\texttt{gulam@atalgo.com} \\
	\And
	{Roshankumar Bisoi}
    \thanks{https://www.linkedin.com/in/rkbisoi777/} 
        \\
	Atalgo Engineering\\
	Atalgo Computing Pty Ltd\\
	Perth, WA, 6000 \\
	\texttt{roshan@atalgo.com} \\
	\And
	{Waliur Rahman}
    \thanks{https://www.linkedin.com/in/waliurrahman42} 
        \\
	Atalgo Engineering\\
	Atalgo Computing Pty Ltd\\
	Perth, WA, 6000 \\
	\texttt{wali@atalgo.com} \\
	\And
	{Dr. Tom Osborn} 
    \thanks{https://www.linkedin.com/in/tomosborn}
    \\
	Atalgo Engineering\\
	Atalgo Computing Pty Ltd\\
	Perth, WA, 6000 \\
	\texttt{tom.osborn@atalgo.com} \\
}

\renewcommand{\shorttitle}{\textit{arXiv} Template}

\hypersetup{
pdftitle={A template for the arxiv style},
pdfsubject={q-bio.NC, q-bio.QM},
pdfauthor={David S.~Hippocampus, Elias D.~Striatum},
pdfkeywords={First keyword, Second keyword, More},
}

\begin{document}
\maketitle

\begin{abstract}
	This paper presents a novel approach to represent enterprise web application structures using Large Language Models (LLMs) to enable intelligent quality engineering at scale. We introduce a hierarchical representation methodology that optimizes the few-shot learning capabilities of LLMs while preserving the complex relationships and interactions within web applications. The approach encompasses five key phases: comprehensive DOM analysis, multi-page synthesis, test suite generation, execution, and result analysis. Our methodology addresses existing challenges around usage of Generative AI techniques in automated software testing by developing a structured format that enables LLMs to understand web application architecture through in-context learning. We evaluated our approach using two distinct web applications: an e-commerce platform (Swag Labs) and a healthcare application (MediBox) which is deployed within Atalgo engineering environment. The results demonstrate success rates of 90\% and 70\%, respectively, in achieving automated testing, with high relevance scores for test cases across multiple evaluation criteria. The findings suggest that our representation approach significantly enhances LLMs' ability to generate contextually relevant test cases and provide better quality assurance overall, while reducing the time and effort required for testing.
\end{abstract}

% keywords can be removed

\section{Introduction}
\label{sec:introduction}
Enterprise web applications constitute the fundamental infrastructure that orchestrates intricate organizational processes and facilitates multiple user interactions fulfilling thee processes \cite{Calero2010ib}. These applications have transcended their traditional role as mere digital interfaces to become critical determinants of operational excellence and market competitiveness for an enterprise. The exponential growth in application complexity, coupled with heightened user expectations, has elevated quality engineering (QE) to a position of paramount significance in the software development lifecycle. This emerging paradigm encompasses a comprehensive framework of methodologies and practices designed to ensure robust functionality, seamless scalability, and sustainable maintainability of enterprise web applications. Within this context, the accurate representation and analysis of architectural intricacies in enterprise web applications has emerged as crucial elements in achieving holistic quality assurance objectives.
When it comes to formal computational model of quality engineering processes, precise and formal structural representation of an enterprise application becomes critical. There is a strong correlation between accurate architectural comprehension and an autonomous system’s ability to provide acceptable level of quality assurance \cite{Marinescu2019rx}. The evolution of enterprise web applications from monolithic structures to complex, distributed systems has created an imperative for more sophisticated methods of automated structural analysis before applying quality assurance processes.
Software testing represents a critical aspect of a software engineering lifecycle, serving as a determinant of system reliability, functional integrity, and overall quality of the software that will eventually support key processes of a business at scale. As contemporary software systems grow increasingly sophisticated, the imperative for comprehensive testing methodologies becomes progressively more pronounced. Beyond the conventional paradigm of defect identification and vulnerability remediation, robust testing frameworks validate system compliance across diverse operational environments while ensuring adherence to specified requirements. This multifaceted approach serves as a crucial safeguard against system failures, security breaches, and compromised user experiences.
The evolution of AI augmented intelligent quality engineering practices transcends mere defect detection, managing the entire lifecycle independently and autonomously \cite{Jeremic2023bn}. 
The recent applications of AI technologies have created significant interest from research as well as business community to explore how a formal and computational model of quality engineering can provide more efficient quality assurance \cite{Kulkarni2024vh}. Recent AI algorithms and technology stack are showing substantial promise to the testing ecosystem through automation capabilities, adaptive learning mechanisms, and predictive analytics. This synergy between AI and testing methodologies has significant potential impact on traditional approaches by optimizing test procedures, minimizing manual intervention, and enabling intelligent test case generation coupled with automated anomaly detection systems.
Natural Language Processing (NLP), a specialized domain within AI, has initiated a transformation in software testing practices by introducing sophisticated linguistic comprehension capabilities to the testing environment \cite{Leotta2024wj}. Generative AI based methodologies enable the formal interpretation of the requirements written in natural language or semi formal format \cite{Bayri2023zs}. This facilitates context-aware testing strategies which are a paradigm shift from traditional testing approaches. By incorporating semantic understanding into the testing framework, new approach enables the intelligent design of test cases based on specifications and user feedback analysis autonomously. This not only expedites the identification of ambiguities and inconsistencies but also frees up the time of the software development team to be able to focus on other areas \cite{Deming2021fy}. 

Nevertheless, as per our research work as part of developing a computational intelligence-based quality engineering platform, we found that the task of effectively capturing and representing the intricate architectural patterns and interaction paradigms within enterprise web applications presents formidable challenges. The heterogeneous nature of development frameworks, technological stacks, and architectural design patterns introduces substantial complexity in modeling application structures with a level of precision we need such systems to work autonomously. Contemporary methodologies frequently demonstrate limitations in providing dynamic and adaptive representations, thus constraining their efficacy in addressing the sophisticated requirements of modern enterprise systems as the system change occurs periodically.  This introduces growing challenges in the quality assurance area as existing script for testing might break and provide false result that might lead to false assumptions and failure to detect crucial bugs in the software systems as the system gets updated periodically \cite{Svacina2021pl}. 
The NLP based solutions for test automation tried to solve this problem with self-healing. Self-healing approaches were effective for small changes in the software system but failed when massive changes were introduced to software systems \cite{Boukhlif2024}. Also, NLP based solutions helped only on test script maintenance but not in initial test script generation or test result reporting. The reason behind was the lack of overall system understanding of the NLP based systems \cite{Fan2023sb}. 
The emergence of Large Language Models (LLMs), particularly the Generative Pre-trained Transformer (GPT) architecture, has presented us with unprecedented opportunities within the software engineering domain \cite{Wang2024gx}. These models exhibit exceptional prowess in natural language processing and demonstrate remarkable capabilities in generating contextually pertinent and semantically enriched content. Such capabilities can be strategically leveraged to autonomously generate comprehensive representations of web application structures through the systematic interpretation of user interactions, technical documentation, and source code artifacts. The integration of LLMs enables the creation of intelligent, adaptive representations that can serve as a robust foundation for quality engineering initiatives, encompassing automated testing frameworks, anomaly detection mechanisms, and performance optimization strategies \cite{Baresi2024wq}.
This research presents a novel methodological framework for representing enterprise web application structures through the strategic deployment of LLMs. Our proposed approach transcends traditional static modeling by not only capturing fundamental elements such as page hierarchies and navigation patterns but also dynamically modeling complex user interactions and system behaviors \cite{Yu2023}. The integration of LLM-driven representation mechanisms into established quality engineering workflows offers organizations substantial opportunities for operational efficiency enhancement and quality assurance optimization.

This paper presents an efficient approach to leveraging Large Language Models for representing enterprise web application structures, specifically focusing on their application in quality engineering \cite{Yu2023}. Our approach addresses several key challenges identified in current literature: the need for dynamic representation of application structure, the ability to capture complex relationships between components, the hierarchical flow of navigation elements and the integration of this representation into existing quality engineering processes. By utilizing LLMs' natural language understanding capabilities, we propose a method that bridges the gap between human understanding and machine-processable representation of web application architectures. Thus, in this research, we have showcased all of the aspects of a software testing process for a web application; test case generation \cite{junior2023casestudytestcase}, test script maintenance and reporting of the test results.
The significance of this research lies in its potential to enhance quality engineering practices by effectively employing generative AI based computational intelligence through improved understanding and representation of enterprise web applications. Our approach not only addresses the limitations of current methods but also provides a foundation for more intelligent and adaptive quality engineering processes in the context of modern web development

\section{Background and Related Work}
\label{sec:background}
Recent years have witnessed growing research interest in converting natural language requirements automatically into functional test scripts, driven by the increasing adoption of agile development and continuous integration practices. This review examines key developments in the field, with a particular focus on systematic analyses, approaches, and automation tools that enable requirements to be transformed into executable test scripts.
The systematic literature review by Mustafa et al. \cite{mustafa} represents seminal work in this domain. Their analysis presents a structured overview of different automated test generation approaches derived from requirements specifications. A key finding emphasizes how test generation approaches must be carefully matched to handle the inherent characteristics of requirements, including their potential ambiguity and incompleteness. Building on this foundation, some researchers \cite{farooqMS} developed an extensive classification system for requirement-based test automation techniques, while also identifying critical research gaps and obstacles that need to be addressed to improve these methods' efficacy. Both research efforts highlight the critical need to develop sophisticated frameworks capable of accurately processing natural language requirements and producing corresponding test scripts. 
Taking a more practical perspective, Chrysalidis et al. developed an innovative semi-automated toolchain system that converts natural language requirements into executable code specifically for flight control platforms \cite{Chrysalidis}. Their work illustrates how domain knowledge can be effectively combined with automation to streamline test script creation. The system enables engineers to structure requirements in modules that can then be automatically transformed into executable tests, providing a concrete implementation of theoretical concepts outlined in systematic reviews.
The research by Koroglu \& Şen  breaks new ground by applying reinforcement learning techniques to generate functional tests from UI test scenarios written in human-friendly languages like Gherkin \cite{Koroglu}. Their approach tackles the complex challenge of converting high-level declarative requirements into concrete test scripts, effectively connecting natural language specifications with automated testing frameworks. Their research suggests machine learning can substantially improve both the precision and efficiency of test script generation. The literature also emphasizes the importance of automation tools like Selenium, with Rusdiansyah outlining optimal practices for web testing using Selenium, highlighting its capabilities for automating user interactions and enhancing test accuracy \cite{Rusdiansyah}. This aligns with Zasornova's observations that automated testing enables improved efficiency and faster detection of defects - crucial factors in contemporary software development \cite{Zasornova}.
Yutia presents an alternative approach through keyword-driven frameworks for automated functional testing. This methodology enables testers to develop succinct, adaptable test cases - particularly valuable when working with evolving natural language requirements \cite{Yutia}. The previous works on automated test script generation demonstrates consistent efforts to advance the script generation from natural language requirements. The combination of systematic reviews, practical toolchains, and sophisticated approaches including reinforcement learning and keyword-driven frameworks shows a comprehensive strategy for addressing industry challenges.
Malik et al. \cite{Malik} made significant contributions through their work on automating test oracles from restricted natural language agile requirements. Their proposed Restricted Natural Language Agile Requirements Testing (ReNaLART) methodology employs structured templates to facilitate test case generation. This research demonstrates how Large Language Models can effectively interpret and transform natural language requirements into functional test scripts, addressing inherent language ambiguities. Additionally, Kıcı et al. \cite{bert} investigated using BERT-based transfer learning to classify software requirements specifications. Their research reveals that Large Language Models can substantially enhance requirements understanding through classification, which subsequently aids in automated test script generation. Their findings emphasize the potential impact of Large Language Models in improving both accuracy and efficiency in test script generation processes. 
Raharjana et al. \cite{Raharjana} conducted a systematic literature review examining how user stories and NLP techniques are utilized in agile software development. Their analysis demonstrates how NLP enhanced by Large Language Models can extract testable requirements from user stories for conversion into automated test scripts, reflecting the increasing adoption of LLMs to connect natural language requirements with automated testing frameworks.
Liu et al. \cite{Liu} introduced MuFBDTester, an innovative mutation-based system for generating test sequences for function block diagram programs. While their work centers on mutation testing, the core principles of generating test sequences from specifications can be enhanced through LLM integration. The ability of LLMs to parse complex specifications enables more efficient test sequence generation aligned with intended software functionality. Complementing this research, explores automated testing challenges in complex software systems, suggesting that LLMs could play a vital role in generating test cases that accurately reflect user requirements and enhance testing processes.
Large Language Models like GPT-3 and its successors have shown exceptional capability in natural language understanding and generation, making them ideal for automated test script creation. Ayenew explores NLP's potential for automated test case generation from software requirements, emphasizing the efficiency benefits of automation \cite{Ayenew}. These findings align with Leotta et al., who demonstrate that NLP-based test automation tools can dramatically reduce test case creation time, making testing more accessible to professionals without extensive programming expertise \cite{Leotta}. Wang et al. present an NLP-driven approach for generating acceptance test cases from use case specifications, showing how recent NLP advances facilitate test scenario identification and formal constraint generation, thereby improving the accuracy of test scripts derived from natural language requirements \cite{Wang}.
Beyond LLMs, researchers have explored various NLP applications in report generation across different domains. Chillakuru et al. examine NLP's role in automating neuroradiology MRI protocols, demonstrating its ability to convert unstructured text into structured reports \cite{Chillakuru}. This capability translates well to software testing, where NLP can extract test scenarios from natural language requirements to streamline report generation. Bae et al. further demonstrate NLP's versatility in automatically extracting quality indicators from free-text reports, a capability that can ensure generated test scripts align with quality standards specified in natural language requirements \cite{BaeJung}.
Similarly, Tignanelli et al. highlight the use of NLP techniques to automate the characterization of treatment appropriateness in emergency medical services, emphasizing the potential for NLP to enhance the quality and relevance of automated reports in various domains \cite{Tignanelli}. Despite these advancements, several research gaps remain in the application of LLMs and NLP techniques for functional test automation. One significant gap is the need for empirical validation of the effectiveness of NLP-based tools compared to traditional testing methods. Leotta et al. note that while many NLP-based tools have been introduced, their superiority has not been rigorously tested in practice \cite{huang2022aeonmethodautomaticevaluation}. 
Additionally, there is a lack of comprehensive frameworks that integrate LLMs with existing testing methodologies, which could facilitate a more seamless transition from natural language requirements to automated test scripts. Furthermore, the challenge of handling ambiguous or incomplete requirements in natural language remains a critical issue. While LLMs have shown promise in interpreting complex text, the variability in natural language can lead to misinterpretations that affect the quality of generated test scripts. Research by Jen-Tse et al. Jen-tse et al. indicates that many generated test cases may not preserve the intended semantic meaning, leading to high false alarm rates. Addressing these challenges through improved training methodologies and more robust NLP techniques is essential for enhancing the reliability of automated test generation \cite{huang2022aeonmethodautomaticevaluation}.

\section{Methodology}
\label{sec:methodology}
Some applications of Large Language Models have been effective in understanding the natural language. However, challenges remain with respect to the amount of data that can be used as a context to leverage the few shot learning of LLM. This makes the usage of LLMs in specialist domains such as test automation quite challenging. One of the solutions is to fine tune the LLM to large amount of data related to a specific field which may result in better reasoning, but then again, it is time consuming and costly process and does not work well for dynamic data. For specific application of enterprise test automation, in context learning is the best approach. This is particularly useful for feeding large web DOM structures to utilize LLM for automation script generation, overall site sense making and to get important insight from the website. After struggling with the limitations of in context learning and trying few approaches such as chunking, we have developed a novel approach to express overall site structure so that the hierarchy remains intact. This is a critical element in our intelligent quality engineering solution. 
This research introduces an innovative methodology to construct a hierarchical structural representation of enterprise web applications, optimized for few-shot learning in large language models (LLMs). The approach leverages state-of-the-art functional test automation principles to ensure scalability, modularity, and enhanced contextual understanding. The proposed methodology is divided into four phases, each targeting a specific aspect of web application analysis and representation.

% \lipsum[4] See Section \ref{sec:headings}.

\subsection{Phase 1: Comprehensive DOM Analysis and Data Structuring}
The first phase involves a comprehensive analysis of the Document Object Model (DOM) of the target web application. Utilizing a custom scraping agent, the methodology extracts all interactive and non-interactive elements from every page of the application, starting from the base URL. Key features of this phase include:
% \lipsum[5]
% \begin{equation}
% 	\xi _{ij}(t)=P(x_{t}=i,x_{t+1}=j|y,v,w;\theta)= {\frac {\alpha _{i}(t)a^{w_t}_{ij}\beta _{j}(t+1)b^{v_{t+1}}_{j}(y_{t+1})}{\sum _{i=1}^{N} \sum _{j=1}^{N} \alpha _{i}(t)a^{w_t}_{ij}\beta _{j}(t+1)b^{v_{t+1}}_{j}(y_{t+1})}}
% \end{equation}
\begin{itemize}
    \item \textbf{Preservation of Hierarchical Context:} Parent-child and sibling relationships among elements are meticulously maintained, along with attributes like locators, URLs, and element types.
    \item \textbf{Semantic Segmentation:} Elements are categorized into logical sections, such as navigation, forms, and content, based on their roles and attributes.
\end{itemize}

The extracted information is encapsulated into structured representations to facilitate downstream processing. This structuring preserves the integrity of the application's element hierarchy and contextual relationships, ensuring compatibility with LLM-based reasoning.

\subsection{Phase 2: Multi-Page Analysis and Site-Wise Synthesis}
The methodology extends individual page analysis to a multi-page context by synthesizing relationships across the application. This phase comprises:

\begin{enumerate}
    \item \textbf{URL Pattern Analysis:} Identifying dynamic segments, common patterns, and page categorizations.
    \item \textbf{Cross-Page Relational Mapping:} Establishing navigation flows, parent-child page hierarchies, and sibling relationships.
    \item \textbf{Contextual Augmentation:} Enhancing page-level summaries with inter-page context and relational metadata.
    \item \textbf{Page Type Identification:} We employed pattern recognition and context aware LLM to find out page types by analysing contents of the pages. The page types are members of a closed set,
    \[
    PT = \{\text{``login'', ``signup'', ``account'', ``listing'', ``detail'', ``form'', ``static''}\}
    \]
    For any page from any web application the type of page \(pt \in PT\).
    \item \textbf{Hierarchical Structure Representation:} To render the extracted data suitable for LLM few-shot learning, the methodology employs a hierarchical encoding mechanism. The hierarchical representation captures:
    \begin{enumerate}
        \item \textbf{Page-Level Summaries:} High-level descriptions of each page, including its type (e.g., login, listing, detail).
        \item \textbf{Section-Level Contexts:} Logical segmentation of elements within a page, annotated with semantic and functional metadata.
        \item \textbf{Element-Level Details:} Attributes, roles, and contextual relationships of individual elements.
        \item \textbf{Navigation Flow:} Our approach analyses and determine the navigation flow and navigation priority based on page content and element attributes
    \end{enumerate}
\end{enumerate}
The outcome of this phase is a comprehensive site structure that encodes navigational pathways, interactive dependencies, and dynamic state transitions and type of pages. This hierarchical structure is crafted to maximize LLM interpretability while minimizing input size constraints. Advanced chunking algorithms ensure that the hierarchical integrity is preserved, enabling effective reasoning over multi-layered representations. Our approach to represent the web application with this structure format maximize outcome by utilizing LLM’s in context learning. As out test generation approach is instruction driven, we can leverage this representation of web application structure to generate test cases which strictly follows instruction.

\subsection{Phase 3: Contextual and Relevant Test Suite Generation and Validation}
The final phase demonstrates the utility of the hierarchical structure through the generation of context-aware test suites. The overall site representation is being passed to LLM in a formatted prompt in an iterative manner for each page. Each prompt contains information about already generated test cases, extracted URL patterns, provided instructions, available elements, required test types. Required test types are combination of predefined test types, E and extracted test types from instruction, A. Required test types 
\[
R = E \cup A
\] 
This involves:
\begin{enumerate}
    \item \textbf{Test Type Determination:} Identifying requisite test types based on the hierarchical representation and page types.
    \item \textbf{LLM-Driven Test Case Creation and Validation:} Employing generative AI to construct detailed test cases that are validated for structural integrity, uniqueness, and contextual relevance with instruction. Relevancy with page flow is also being checked under test case validation. Test cases returned by LLM are being incorporated with structured format for scalability and it makes validation easier. Based on the information regarding page and navigation priority that are being passed as site structure information to LLM, LLM can predict the test case priority. Based on our evaluation and testing so far, we have found the information received from LLM after this operation to be correct approximately 87\% of the time.
    \item \textbf{Iterative Refinement:} Incorporating feedback loops to optimize test quality, ensuring alignment with application behaviour and hierarchical context. 
\end{enumerate}
The generated test suites serve as a practical validation of the methodology's capability to encapsulate enterprise web application structures in a format conducive to few-shot learning and automated reasoning. 

\subsection{Phase 4: Test Suite Execution}
After we have generated a semantically rich formal representation of the web application in four phases above, we ca execute test cases by driving a test framework. We also need to map specific test data to test steps under each test cases or alternatively generate synthetic data to execute autonomous test cases. This phase has two major parts:
\begin{enumerate}
    \item \textbf{Test Data Mapping or Synthetic Data generation:} Based on element pattern and test case, we employ LLM to map test data. The test data schema and test case with valid test steps are used as context for LLM to return proper mapping. If test data is not present, LLM’s in context learning is being used to generate meaningful synthetic data for given iteration to enable test suites for execution.
    \item \textbf{Interpreter for Test Suite to Test Automation Framework Language:} Test automation framework library or tool can be used by writing code in supported programming languages or by using tool’s own language or representations. Our approach involves training LLM with appropriate tool specific knowledge to overcome this interpretation challenge. According to our experiments, LLM tends to understand codes better than tool specific languages. For interpretation, our approach converts each test steps into tool specific representations. The representation uses set of defined actions for the tool to be used. 
    
\end{enumerate}

\subsection{Phase 5: Test Report and Analysis}
All the executed test suites by the proposed system produce results. The results produced by the system can be easily understandable by test engineers but might not make much sense to individuals from non testing background. Thus our approach involves LLM based summaries on top of the result that is easy to understand to any individual with a bit of technical knowledge. Our representation of result also leverages LLM’s in context learning as the result must co relate with the web application on which the test cases got generated in the first place. To point out exact issue of the system from the test result, our representation of the web application again plays a role as a test report enhancer. This behaviour mimics experienced test engineers who has extensive knowledge about the web application while producing a reporting about test results. 

The entire phases are represented in Fig 1. Figure 1 illustrated the process and data flow of the phases. Phase one starts with scraper, then phase 2 is about formatting the data to the proposed representation. From phase 3 the formatted data starts getting utilized and on phase 4 the represented format gets expanded. Finally on phase 5, the final reasoning step, through complete test reporting, the process flow ends. 

\begin{figure}
    \centering
    \includegraphics[width=0.8\textwidth]
    {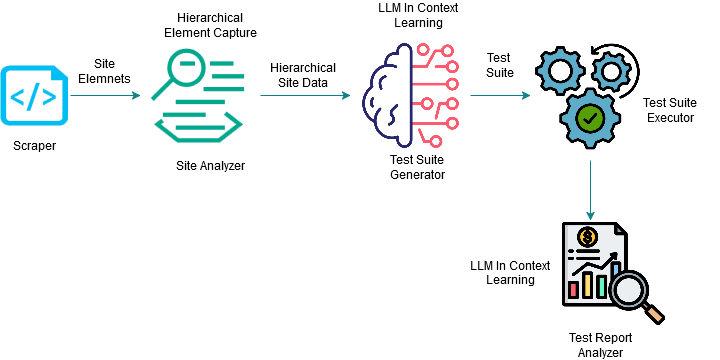} % Adjust width as needed
    \caption{Overall phase flow}
    \label{fig:fig1}
\end{figure}

\section{Experiment}
\label{sec:experiment}

To evaluate our novel approach for developing hierarchical structure representation of enterprise web applications and its utility in enabling few-shot learning with large language models (LLMs), we conducted comprehensive experiments. These experiments were designed to demonstrate the effectiveness of AI-driven methodologies, including web scraping, site analysis, automated test case generation, and execution, across different web application contexts.

\subsection{Objectives}                                                    
\begin{enumerate}
    \item Validation of scraped data for sense making from web applications
    \item Validation of hierarchical representation capability of Document Object Model (DOM) structure and interactive behaviour of web applications.
    \item To assess the automated test case generation process using LLMs 
    \item Assess and validate automated test case generation process using LLMs, focusing on functional validation of web application features
    \item Analyse the effectiveness of few-shot learning in generating meaningful and relevant test cases with minimal human intervention.

\end{enumerate}

\subsubsection{Experimental Setup}
The experiments were carried out in two different enterprise Web applications:
\begin{enumerate}
    \item \textbf{Swag Labs:} An \hyphenation{e-commerce} platform, specifically focused on its login functionality hosted at \url{https://www.saucedemo.com/}.
    \item \textbf{MediBox} A \hyphenation{healthcare-focused}  application, with emphasis on user signup functionality. The application is hosted in Atalgo Engineering development environment.

\end{enumerate}
Both applications provided an opportunity to test our approach under different structural and functional paradigms, allowing a broad evaluation of the methodology.

\subsection{Scope of Experiments}
To evaluate our approach, we need to evaluate results of each phase individually. Each phase depends on our data representation, and we have experimented each phase with our data representation to assess the correct representation. In this section, we will discuss the experiments conducted.
\subsubsection{Web Scraping and Data Extraction}
For both applications, we utilized an AI-powered scraping module to extract metadata for each HTML element. This process was essential for building a structured hierarchical representation of the web applications. We have tried extracting all features and even tried extracting and passing the entire DOM tree. The earlier approaches induced problems. There were lot of unnecessary data which was clouding LLM’s decision and resulting in a number of irrelevant test cases. Also, the results of web application sense making through LLM was not up to the mark. Through our experiments, we found that elements need unique identifier apart from their locator. Locators for sibling elements look almost similar sometimes, which can potentially mislead LLM and generated test case might end up with wrong elements. The overall DOM tree can be large which makes it impractical to provide overall data to LLM for web application sense making through in context learning as LLMs have fixed context window. We cannot feed LLM with the entire DOM tree, even essential extracted features only are impractical. We have to make sure our solution works for any web application as this is going to be the foundational technology behind generative AI based intelligent quality engineering. For the reasons outlined above, traditional chunking method also does not work which we tried before developing this approach. In our current approach, we preserved the hierarchical order and then chunked DOM tree page wise. We captured navigation flow as well. That way even after page wise DOM tree chunking ensured LLM’s understanding of the page flow and the next steps for test generation.  The extracted data included:
\begin{enumerate}
    \item \textbf{Element Identification:} Each element was assigned a unique identifier. For our experiment we have observed after adding element identifier we got rich test cases for both Medibox and SauceLab applications. LLM had clear understanding and perfect mapping of elements for signup form and signin form respectively.
    \item \textbf{Hierarchy and Structure:} Parent-child relationships were captured to map the DOM structure. We have extracted page URL flow, element flow and URL relationship. Experimental data shows, LLM understands element flow and page hierarchy for actions. Which led LLM to successfully generate multi page or section test cases.
    \item \textbf{Attributes and Content:} Metadata such as tag names, attributes (e.g., href, type, onclick), and textual content (e.g., placeholders, labels) were documented. These attributes and contents help LLM to understand about elements functionality with more clarity
    \item \textbf{Geometric and Visual Data:} Element dimensions (width, height) and positions (x, y) were extracted to support usability analysis. Visual and geometric features of elements are crucial for later on steps to identify the elements locator and elements hierarchy, in short, the order of the elements
    \item \textbf{Interaction Properties:} Visibility and interactivity of elements were flagged for validation during test execution. During validation, rule-based validation remains successful, because of early element property identification.
\end{enumerate}
For instance, in Swag Labs, interactive elements such as the username and password input fields, login button, and error containers were captured. Similarly, in MediBox test application, key elements on the signup form, including fields for full name, contact number, email, password, and confirmation password, were identified.

\subsubsection{Site Analysis and Hierarchical Representation}
The scraped data was fed into a site analysis module to organize elements into logical sections and establish their relationships. This module created a hierarchical representation of the web applications by:
\begin{itemize}
    \item Categorizing elements into sections such as navigation links, input forms, and feedback mechanisms.
    \item Mapping relationships to construct a complete DOM tree, allowing downstream processing to focus on key interactive regions of the applications.
\end{itemize}
The analysis facilitated by our approach effectively transformed unstructured web data into a hierarchical format, serving as a foundation for automated test case generation.

\subsubsection{Automated Test Case Generation Using LLMs}
Our methodology leveraged generative AI and LLMs to automatically generate functional test cases. The structured hierarchical data served as input, and the LLMs were prompted to create test cases covering diverse scenarios, including:
\begin{enumerate}
    \item \textbf{Field Validation:} Ensuring the presence and functionality of input fields.
    \item \textbf{Error Handling:} Validating error messages for invalid or missing inputs.
    \item \textbf{Navigation and Redirection:} Verifying transitions between application pages.
    \item \textbf{Data Consistency:} Ensuring logical consistency between dependent fields, such as password and confirmation password.
\end{enumerate}
For example, in MediBox, test cases included verifying unique email and mobile number registration, password strength validation, and successful user registration.

\subsubsection{Test Case Execution and Data Handling}
Once generated, the test cases were executed systematically against the respective applications. The execution process included:
\begin{itemize}
    \item Simulating real-world interactions with UI elements.
    \item Validating functionality such as field visibility, error message accuracy, and navigation flows.
    \item Handling input data through synthetic data generation to cover edge cases and maintain flexibility in testing.

\end{itemize}
Execution results, including logs and screenshots, were stored for further analysis.

\subsection{Applications and Scope}
The hierarchical representation enabled effective utilization of LLMs in a few-shot learning setup, reducing the dependency on large labelled datasets. By focusing on structural and functional representations of web applications, the approach provided the following advantages:
\begin{itemize}
    \item Scalability: The pipeline could adapt to various web applications with minimal customization.
    \item Automation: From data extraction to test case execution, the process minimized human intervention.
    \item Precision: The hierarchical representation ensured that LLMs received contextually relevant prompts, leading to accurate and comprehensive test cases.

\end{itemize}

\subsection{Experimental Design Highlights}
For Swag Labs, 10 functional test cases focused on login validation were developed and executed. For MediBox, another set of 10 test cases targeted the user signup process, ensuring robust field validation and error handling. Each test case was categorized by priority, with high-priority cases addressing core functionalities such as login authentication and secure user registration.

\subsection{Execution Workflow}
The execution of test cases across both Swag Labs and MediBox applications followed a systematic workflow, leveraging the hierarchical representation to guide the interactions:
\begin{enumerate}
\item \textbf{Test Case Initialization:}
\begin{itemize}
\item Generated test cases were categorized based on priority and functional grouping.
\item Each test case contained a detailed sequence of actions, expected outcomes, and validation criteria.
\end{itemize}
\item \textbf{Environment Setup:}
\begin{itemize}
\item A controlled testing environment was configured for both applications, ensuring consistent and reproducible results.
\item Testing scripts were deployed to interact directly with the DOM elements as identified in the hierarchical structure.
\end{itemize}
\item \textbf{Simulation of User Interactions:}
\begin{itemize}
\item The execution engine simulated user interactions, such as entering text in input fields, clicking buttons, and navigating between pages.
\item For Swag Labs, the interactions included logging in with various credential combinations (e.g., valid, invalid, empty fields).
\item For MediBox, the tests involved filling out the signup form with different datasets to validate field-level constraints and consistency.
\end{itemize}
\item \textbf{Real-Time Validation:}
\begin{itemize}
\item Each interaction was validated against the expected behaviour.
\item Success and failure conditions were logged, with detailed error messages captured for deviations.
\end{itemize}
\item \textbf{Dynamic Data Handling:}
\begin{itemize}
\item Synthetic data generation modules were employed to provide realistic test inputs, particularly for MediBox's signup form, covering edge cases like invalid email formats and weak passwords.
\item This ensured that the testing covered a wide spectrum of potential user inputs.
\end{itemize}
\item \textbf{Logging and Reporting:}
\begin{itemize}
\item Comprehensive logs, including timestamps, actions performed, and results, were generated.
\item Screenshots were captured for failed test cases to aid in debugging and root cause analysis.
\end{itemize}
\end{enumerate}

\subsection{Challenges Encountered}
While the experimental pipeline was robust, certain challenges highlighted areas for further refinement:
\begin{enumerate}
\item \textbf{Element Accessibility:}
\begin{itemize}
\item Some UI elements required additional efforts for identification due to dynamic CSS selectors or hidden states, particularly in MediBox's signup flow.
\end{itemize}
\item \textbf{Latency in Response Validation:}
\begin{itemize}
\item Delays in server responses caused intermittent issues during real-time validation, necessitating timeout adjustments.
\end{itemize}
\item \textbf{Dynamic Data Dependencies:}
\begin{itemize}
\item Handling dependencies between fields (e.g., password and confirmation password) required specific logic to ensure consistent behaviour.
\end{itemize}
\end{enumerate}

\subsection{Evaluation Metrics}
To evaluate the effectiveness of the experiment, the following metrics were used:
\begin{enumerate}
\item \textbf{Coverage:} The proportion of application functionalities covered by generated test cases.
\item \textbf{Execution Success Rate:} The percentage of test cases successfully executed without errors.
\item \textbf{Error Categorization:} Classification of failures to identify patterns and prioritize fixes (e.g., UI-related, server-side issues).
\item \textbf{Efficiency:} Time taken per test case execution and overall testing cycle duration.
\end{enumerate}

\section{Result and Analysis}
\label{sec:result}
As part of our experiments, we have generated and executed test suites. The ultimate goal is to be able to provide quality assurance on the software application using generative AI. For the purposes of this specific research, we wanted to confirm whether our unique approach of site representation for the purposes of enabling LLM to conduct in context learning helps us achieve effective quality engineering. We have discussed as part pf methodology how this approach of web application representation is foundational to achieve AI augmented quality engineering. To evaluate the overall efficiency of our data structure, we have assessed the quality of the generated test cases. The evaluation criteria we established for this assessment are:

\begin{enumerate}
    \item Test case execution success rate
    \item  Relevance of the test cases based on instruction provided
    \item Relevance of the test cases for the web application under test
    \item Relevance and accuracy of test data mappings
    \item Quality of synthetic data generation

\end{enumerate}
These criteria for evaluation are relevant for any enterprise test automation project and are regularly used by automation engineers, implicitly or explicitly. The final test execution results are produced by the Selenium test automation tool which is orchestrated by our AI platform, Flame.
Based on the outlined evaluation criteria, the performance and efficiency of the developed approach for representing enterprise web application structures using Large Language Models (LLMs) in the service of intelligent quality engineering were analysed using two distinct use cases: the Swag Labs web application and the Atalgo Engineering’s development/testing environment, known as MediBox platform. We present below results of various components of our experiments, highlighting the relevance and impact of the novel structure in overall functional testing. We will show case the generated test suites for both MediBox and Swag Labs platforms then we will provide assessment against the evaluation criteria.

\begin{table}
	\caption{Generated Test Suite for Medibox Application}
	\centering
	\begin{tabularx}{\textwidth}{lXlX}
		\toprule
		\textbf{Test Case ID} & \textbf{Test Case Name} & \textbf{Priority} & \textbf{Description} \\
		\midrule
		TC01 & Verify navigation to User Signup page & High & Check if the User Signup page is navigable from the homepage. \\
		TC02 & Verify unique mobile number registration & High & Ensure that the mobile number is unique for each user during the registration process. \\
		TC03 & Verify unique email address registration & High & Validate that the email address is unique for each user during registration. \\
		TC04 & Verify password and confirm password match & High & Check if the password and confirm password fields match before form submission. \\
		TC05 & Verify required fields on User Signup form & High & Test if Email field accepts valid input. \\
		TC06 & Verify successful user registration & High & Ensure all required fields must be filled before form submission. \\
		TC07 & Verify password strength requirement & Medium & Ensure the password meets the required strength criteria. \\
		TC08 & Verify email address format validation & High & Check if the email address entered follows the correct format. \\
		TC09 & Verify mobile number format validation & Medium & Check if the mobile number entered follows the correct format. \\
		TC10 & Verify navigation to Sign In page after registration & High & Check if the user can navigate to the Sign In page after registration. \\
		\bottomrule
	\end{tabularx}
	\label{tab:test_suite}
\end{table}

\begin{table}
	\caption{Generated Test Suite for Swag Labs}
	\centering
	\begin{tabularx}{\textwidth}{lXlX}
		\toprule
		\textbf{Test Case ID} & \textbf{Test Case Name} & \textbf{Priority} & \textbf{Description} \\
		\midrule
		TC01 & Login with valid credentials & High & Test login functionality using valid username/password. \\
		TC02 & Login with invalid username & High & Test login with invalid username and valid password. \\
		TC03 & Login with invalid password & High & Test login with valid username and invalid password. \\
		TC04 & Login with empty username & Medium & Validate login error when username is empty. \\
		TC05 & Login with empty password & Medium & Validate login error when password is empty. \\
		TC06 & Login with locked-out user & Low & Test locked-out user credentials return proper error. \\
		TC07 & Login with performance glitch user & Medium & Test login using performance\_glitch\_user. \\
		TC08 & Login with all fields empty & Low & Validate error messages for empty username/password. \\
		TC09 & Verify error message for invalid input & Medium & Validate error message when credentials are invalid. \\
		TC10 & Login with problem user & Low & Test login using problem\_user and validate issues. \\
		\bottomrule
	\end{tabularx}
	\label{tab:swag_labs_test_suite}
\end{table}

\subsection{Evaluation}
We have split the entire evaluation process into 5 steps. As the novel representation method is being used in 5 phases, we had to evaluate each phase to get complete evaluation. The process or phases are sequentially dependent i.e the next step always depends upon the output of previous step. Let's discuss the evaluation steps in this section:

\subsubsection{Test Case Execution Success Rate}
The test execution success rate was calculated as the percentage of test cases that executed successfully without errors (and they were expected to behave the same way). We have used Selenium (Test Automation Platform) to perform automated test execution from the generated test suites. The results for both applications are summarized in Table 3.

\begin{table}[!ht]
	\caption{Test Case Execution Success Rate}
	\centering
	\begin{tabularx}{\textwidth}{lXXXX}
		\toprule
		\textbf{Application} & \textbf{Total Test Cases} & \textbf{Passed Test Cases} & \textbf{Failed Test Cases} & \textbf{Success Rate (\%)} \\
		\midrule
		Swag Labs & 10 & 9 & 1 & 90.00 \\
		MediBox & 10 & 7 & 3 & 70.00 \\
		\bottomrule
	\end{tabularx}
	\label{tab:test_case_success_rate}
\end{table}

\textbf{Swag Labs Results:}

\begin{itemize}
\item \textbf{Total Tests Executed:} 10
\item \textbf{Success Rate:} 90%
\item \textbf{Observations:}
\begin{itemize}
\item 9 out of 10 test cases executed successfully, covering scenarios like valid and invalid login attempts, empty fields, and performance glitch simulation.
\item \textbf{Failure Case:} "Login with performance glitch user" encountered timeout issues, indicating limitations in handling extreme scenarios.
\end{itemize}
\end{itemize}

\textbf{MediBox Results:}

\begin{itemize}
\item \textbf{Total Tests Executed:} 10
\item \textbf{Success Rate:} 70%
\item \textbf{Observations:}
\begin{itemize}
\item 7 test cases passed successfully, including critical validations for email format, mobile number uniqueness, and password matching.
\item \textbf{Failure Cases:}
\begin{itemize}
\item "Navigation to User Signup Page" failed due to inaccessible UI elements.
\item "Successful User Registration" encountered NoneType errors related to delayed responses.
\item "Navigation to Sign In Page After Registration" faced assertion issues for page titles.
\end{itemize}
\end{itemize}
\end{itemize}

\subsubsection{Relevance of Test Cases Based on Instructions}
\textbf{Swag Labs:}
\begin{itemize}
    \item \textbf{Instructions:} Fetch all input fields, buttons, and labels from the homepage of Sauce Demo (\url{https://www.saucedemo.com/}) and provide the details for each element, including their id, class, and attributes. Generate a detailed test plan for logging in to the Sauce Demo application using valid credentials (standard\_user/secret\_sauce) and invalid credentials. Include preconditions, steps, expected outcomes, and priority for each test. Create a Selenium test script in Python to automate the login functionality of Sauce Demo. The script should test both successful and failed login attempts and validate error messages. Execute the generated Selenium test script for Sauce Demo and provide a detailed test summary, including pass/fail results, screenshots of failures, and execution time for each test.
    \item Test cases directly addressed the login functionality's critical scenarios, confirming adherence to instructional goals.
    \item Coverage included variations in credentials, field emptiness, and user-specific conditions.

\end{itemize}

\textbf{Medibox:}

\begin{itemize}
\item \textbf{Instructions:} Create and execute a minimum of 10 functional test scripts specifically for the user signup process.
\item \textbf{Deployed on our experiment environment} 
\item \textbf{User Signup Endpoint URL:} /UserSignup
\item Ensure the test cases cover a wide range of scenarios for comprehensive validation.
\end{itemize}

\textbf{Details:}
\begin{itemize}
\item Test cases were comprehensive, targeting user registration with field validations, navigation checks, and logical consistency.
\item Generated scenarios matched the specified instructions for ensuring accurate and secure user onboarding.
\end{itemize}

\subsubsection{Relevance of Test Cases for the Web Application}
\textbf{Swag Labs:}
\begin{itemize}
\item Structural insights from the DOM allowed the generation of contextually relevant test cases tailored to the application's architecture.
\item \textbf{Example:} Dynamic field validations and error feedback mechanisms were precisely targeted.
\end{itemize}

\textbf{MediBox:}
\begin{itemize}
\item Detailed analysis of navigation links and form sections ensured alignment with MediBox’s hierarchical structure.
\item \textbf{Key observation:} Integration of URL pattern recognition enabled mapping of related endpoints, enhancing navigation validation.
\end{itemize}

\subsubsection{Test Data Mappings Relevance}
\textbf{Swag Labs:}
\begin{itemize}
\item Mapped data consistently reflected the application's structural metadata, enabling accurate input-output scenarios.
\item \textbf{Example:} User-specific identifiers like "locked-out user" and "problem user" ensured realistic validation.
\end{itemize}

\textbf{MediBox:}
\begin{itemize}
\item Synthetic data generation effectively complemented user-provided data to fill missing fields during execution.
\item Data mappings accurately represented user inputs, confirming seamless integration of synthetic and real data.
\end{itemize}

\subsubsection{Synthetic Data Generation Contextual Relevance}
\textbf{Swag Labs:}
\begin{itemize}
\item Synthetic data aligned with expected field formats, enabling smooth execution of test cases with varying credentials.
\item \textbf{Observations:}
\begin{itemize}
\item Data diversity enhanced robustness in edge case validations.
\end{itemize}
\end{itemize}

\textbf{MediBox:}
\begin{itemize}
\item Contextual relevance was evident in generated data for email, mobile number, and password fields.
\item \textbf{Observations:}
\begin{itemize}
\item Password strength validation scenarios benefited significantly from the synthesized data.
\end{itemize}
\end{itemize}

\begin{table}[!ht]
    \caption{Summary Table of Results}
    \centering
    \begin{tabularx}{\textwidth}{lXX}
        \toprule
        \textbf{Criteria} & \textbf{Swag Labs Success Rate} & \textbf{MediBox Success Rate} \\
        \midrule
        Test Case Execution Success Rate & 90\% & 70\% \\
        Instruction Relevance & High & High \\
        Web Application Relevance & High & High \\
        Data Mappings Relevance & High & High \\
        Synthetic Data Contextuality & High & High \\
        \bottomrule
    \end{tabularx}
    \label{tab:summary_table}
\end{table}

Although not directly related to the objective of this specific research, we have timeboxed the time taken in each of these activities because in actual real-world projects, speed of execution is of high importance. This provides significant savings at scale. We have concluded that once the initial setup and the configuration of the requirements, instructions etc are complete, this approach provides significant time savings (upwards of 50\%) compared to a traditional test automation approach. This saving is more pronounced in the maintenance phase of the project and becomes significant as the software application scales.
\section{Future Work and Conclusion}
\label{sec:conclusion}
Our research demonstrates the successful application of LLM's few-shot learning capabilities in automated test script generation and execution, despite using models not specifically trained for the functional testing domain. The results validate our approach to hierarchical web application representation while also highlighting areas for future enhancement.
From this point onward, towards the goals of computational model of quality engineering, two primary directions emerge for the future research. First, we plan to fine-tune LLMs specifically for the test automation domain using curated datasets. This specialized training should address current limitations in handling large context windows and input sizes for individual requests. Second, we propose developing a knowledge graph architecture to store and efficiently retrieve element-level data during run-time, potentially reducing the input size required for page-level test case generation.
Although our solution demonstrates robust performance across most web applications in generating quality test suites, we acknowledge certain limitations. Applications with exceptionally large DOM structures can challenge our algorithm's ability to maintain hierarchical relationships. Furthermore, the substantial input sizes required by our representation method may lead to increased costs when using commercial LLM services, although the benefits of improved test coverage and maintenance efficiency will often justify this investment.
The methodology's strength lies in its ability to enable LLMs to comprehend and interact with web application structures through in-context learning, producing contextually relevant test cases. Looking ahead, we believe that the integration of domain-specific fine-tuning and knowledge graph-based element retrieval will significantly mitigate current limitations, further enhancing the scalability and cost-effectiveness of our approach. These improvements will pave the way for more efficient and autonomous quality engineering processes in enterprise web applications.

\bibliographystyle{unsrt}
% \bibliography{references}  %%% Uncomment this line and comment out the ``thebibliography'' section below to use the external .bib file (using bibtex) .

%% Uncomment this section and comment out the \bibliography{references} line above to use inline references.

% 	\bibitem{}



% \end{thebibliography}

\end{document}